\documentclass{article}

\usepackage{PRIMEarxiv}

\usepackage[utf8]{inputenc} % allow utf-8 input
\usepackage[T1]{fontenc}    % use 8-bit T1 fonts
\usepackage{hyperref}       % hyperlinks
\usepackage{url}            % simple URL typesetting
\usepackage{booktabs}       % professional-quality tables
\usepackage{amsfonts}       % blackboard math symbols
\usepackage{nicefrac}       % compact symbols for 1/2, etc.
\usepackage{microtype}      % microtypography
\usepackage{lipsum}         % dummy text
\usepackage{fancyhdr}       % header
\usepackage{graphicx}       % graphics
\graphicspath{{images/}}    % MIGRATED: Set graphics path from content paper

\usepackage[compress]{cite}
\usepackage{textcomp}
\usepackage[table]{xcolor}
\usepackage{subcaption}     % Replaced subfig/subfigure for modern compatibility
\usepackage{array,multirow}
\usepackage{amsmath,amssymb,mathrsfs}
\usepackage[font=small,labelfont=bf]{caption}
\usepackage{algorithm}
\usepackage{algorithmicx}
\usepackage{algpseudocode}
\usepackage{bbm}
\usepackage{bm}
\usepackage{adjustbox}
\usepackage{siunitx}        % for aligning numbers
\usepackage{lmodern}

\newcommand{\goodchange}[1]{\textcolor{red}{#1}}
\newcommand{\badchange}[1]{\textcolor{blue}{#1}}
\title{H2-Cache: A Novel Hierarchical Dual-Stage Cache for High-Performance Acceleration of Generative Diffusion Models}

\author{
    Mingyu~Sung, 
    Il-Min~Kim, 
    Sangseok~Yun, 
    and~Jae-Mo~Kang%
    \thanks{
        Mingyu Sung and Jae-Mo Kang are with the Department of Artificial Intelligence, Kyungpook National University, Daegu, South Korea (Corresponding author: Jae-Mo Kang, e-mail: jmkang@knu.ac.kr).
    }% <- 이 % 기호가 중요합니다.
    \thanks{
        Il-Min Kim is with the Department of Electrical and Computer Engineering, Queen's University, Kingston, K7L 3N6, Canada.
    }%
    \thanks{
        Sangseok Yun is with the Department of Information and Communications Engineering, Pukyong National University, Busan 48513, South Korea (Corresponding author: Sangseok Yun, e-mail: ssyun@pknu.ac.kr).
    }
}

\begin{document}
\maketitle

\begin{abstract}
Diffusion models have emerged as state-of-the-art in image generation, but their practical deployment is hindered by the significant computational cost of their iterative denoising process. While existing caching techniques can accelerate inference, they often create a challenging trade-off between speed and fidelity, suffering from quality degradation and high computational overhead. To address these limitations, we introduce \textbf{H2-cache}, a novel hierarchical caching mechanism designed for modern generative diffusion model architectures. Our method is founded on the key insight that the denoising process can be functionally separated into a structure-defining stage and a detail-refining stage. H2-cache leverages this by employing a dual-threshold system, using independent thresholds ($\tau_{1}, \tau_{2}$) to selectively cache each stage. To ensure the efficiency of our dual-check approach, we introduce pooled feature summarization (PFS), a lightweight technique for robust and fast similarity estimation. Extensive experiments on the Flux architecture demonstrate that H2-cache achieves significant acceleration—up to 5.08x—while maintaining image quality nearly identical to the baseline, quantitatively and qualitatively outperforming existing caching methods. Our work presents a robust and practical solution that effectively resolves the speed-quality dilemma, significantly lowering the barrier for the real-world application of high-fidelity diffusion models. Source code is available at \url{https://github.com/Bluear7878/H2-cache-A-Hierarchical-Dual-Stage-Cache}.
\end{abstract}

% keywords can be removed

\section{INTRODUCTION}
Diffusion models have achieved state-of-the-art (SOTA) performance in image generation \cite{sohl2015deep,song2021score,croitoru2023diffusion}. Their ability to synthesize high-fidelity and diverse images has led to the development of numerous real-world applications \cite{ha2025generative,saharia2022photorealistic}. The underlying principle of these models involves a forward process that gradually noises an image, and a neural‑network‑guided reverse process that removes the noise to recover a clean image. This iterative refinement process is key to their generative capabilities \cite{ho2020denoising}. However, the highly iterative nature of the denoising process imposes a significant computational burden \cite{song2021denoisingimplicit,salimans2022progressive,karras2022elucidating}. This heavy computation leads to high latency, which limits deployment in real‑time settings \cite{rombach2022high}. The heavy processing power these models demand still limits their wider use.

To address these limitations, researchers have explored various methods to accelerate the inference process of diffusion models. Among these, techniques that reuse computations from previous steps have shown considerable promise \cite{lyu2022accelerating}. One notable method, \textit{Block Caching} \cite{wimbauer2024cache} exploits the fact that consecutive denoising steps produce very similar intermediate features. The core idea behind block caching is to cache the output of whole layer blocks (e.g., residual blocks in a UNet backbone) at a specific time step and reuse these cached outputs in subsequent steps. By identifying that the outputs of many layer blocks change minimally from one step to the next, block caching can skip a significant amount of redundant computation, thereby speeding up the generation process.

However, skipping whole blocks comes with important trade‑offs. The aggressive simplification of skipping entire blocks can lead to a loss of high-frequency details, making it harder for the model to apply delicate refinements to the generated image. Simultaneously, the process of checking for cache hits at each block introduces a significant computational overhead. This overhead can sometimes cancel out the speed gains—or even slow the model down overall compared to the original model, especially when using fewer denoising steps.

To solve both issues, we introduce \textbf{H2‑Cache}—a new caching method that keeps images sharp and makes diffusion models run faster. Our approach aims to mitigate the problems of detail loss and high overhead associated with existing block caching techniques.

\subsection{Contributions}
In this paper, we introduce H2-cache, a novel caching framework designed to significantly accelerate modern diffusion model inference while preserving high-fidelity output. Our primary contributions are as follows:

\begin{itemize}
\item We identify and leverage a \textbf{functional dichotomy} within the denoising network, separating its computation into a structure-defining stage ($\mathcal{B}_{L1}$) and a detail-refining stage ($\mathcal{B}_{L2}$). Based on this insight, we propose \textbf{H2-cache}, a novel \textbf{hierarchical, two-stage caching mechanism} that applies independent thresholds ($\tau_{1}, \tau_{2}$) to each stage. This provides granular control over the speed-quality trade-off, a significant advancement over monolithic caching approaches.

\item To make our dual-check strategy computationally feasible, we introduce \textbf{Pooled Feature Summarization (PFS)}, a lightweight yet robust technique for efficient similarity estimation between high-dimensional tensors. PFS serves as a fast, low-overhead proxy for semantic difference, which is critical for enabling our frequent caching checks without negating performance gains.

\item Through extensive experiments on the modern \textbf{Flux} architecture, we demonstrate that H2-cache achieves significant acceleration (up to \textbf{5.08x}) while maintaining near-baseline image quality. Our method substantially outperforms existing caching strategies like block cache and TeaCache in both quantitative metrics (e.g., CLIP-IQA) and perceptual quality.
\end{itemize}

\section{Related Work}
\subsection{Diffusion Models}
Denoising diffusion probabilistic models (DDPMs) \cite{ho2020denoising} have established themselves as a powerful class of generative models. They define a forward process that incrementally adds gaussian noise to an input sample and a corresponding reverse process that learns to denoise the data, starting from pure noise, to generate a sample. The reverse process is typically parameterized by a neural network—typically a U‑Net, which is trained to predict the noise added at each step. Subsequent works, such as denoising diffusion implicit models (DDIM) \cite{song2021denoisingimplicit}, proposed a non-Markovian forward-reverse process, allowing for deterministic sampling and much faster sampling with far fewer steps, though sometimes at the cost of image quality. These foundational models have laid the groundwork for the broad success of diffusion-based image synthesis.

The fusion of diffusion models with natural language understanding has given rise to SOTA text-to-image generation systems. Models like GLIDE \cite{nichol2022glide} demonstrated that guiding a diffusion model with text embeddings can produce photorealistic images that align with complex textual descriptions. This was further advanced by models such as DALL-E 2 \cite{ramesh2022hierarchical}, Imagen \cite{saharia2022photorealistic}, and Stable Diffusion \cite{rombach2022high}, which leverage powerful pre-trained vision-language models (VLMs) such as CLIP \cite{radford2021learningtransferablevisualmodels} to map text and images into a shared embedding space. This allows for robust and nuanced control over the image generation process. The text prompt conditions each denoising step so the final image aligns semantically with the prompt.

\subsection{Block Cache for Diffusion Models}
To reduce the computational redundancy inherent in the iterative denoising process, caching strategies have been introduced. The concept of a block cache \cite{wimbauer2024cache} is a notable recent advancement in this area. This technique is predicated on the observation that the outputs of intermediate blocks within the U-Net architecture change minimally between consecutive time steps. The block cache method stores the output tensors of specific blocks (e.g., ResNet blocks or attention blocks) at a given time step. In the subsequent step, instead of re-computing the block, a quick similarity check tests whether the block’s input has changed significantly. If the change is below a predefined L2-norm threshold, $|\Delta|_2<\tau$, the cached output is reused, effectively skipping a large portion of the model's computation.

While this can lead to substantial speedups, its efficacy is highly dependent on the careful tuning of the cache-hit threshold $\tau$, and it risks quality degradation from overly aggressive caching. Addressing these limitations, TeaCache \cite{liu2025timestep} proposed a learnable caching framework. Instead of a fixed, heuristic-based threshold, TeaCache employs a lightweight policy (teacher) network that learns to predict which blocks are safe to cache at each step. This approach aims to dynamically maximize acceleration while preserving image quality. However, this introduces its own set of challenges, including the overhead of training the separate policy network and potential generalization issues across diverse datasets and prompts. Hence, there remains a need for a method that speeds up diffusion models while keeping quality high—without training an extra helper network.

\section{Preliminaries}
\subsection{Denoising Diffusion Models}
To efficiently model high-resolution images, our approach is based on a latent diffusion model (LDM). An image $\mathbf{x}$ is first mapped into a compact latent vector $\mathbf{z}_0 = \mathcal{E}(\mathbf{x})$ using the encoder of a pre-trained variational autoencoder (VAE). The denoising diffusion process is then applied directly within this latent space.

The process consists of two main parts: a fixed forward diffusion process and a learned reverse denoising process. The forward process, $q$, gradually adds Gaussian noise to an initial latent vector $\mathbf{z}_0$ over a sequence of $T$ time steps:
\begin{equation}
    q(\mathbf{z}_t | \mathbf{z}_{t-1}) = \mathcal{N}(\mathbf{z}_t; \sqrt{1 - \beta_t}\mathbf{z}_{t-1}, \beta_t\mathbf{I}),
\end{equation}
where $\{\beta_t\}_{t=1}^T$ is a predefined variance schedule. A key property of this process is that we can sample $\mathbf{z}_t$ at any time step $t$ in a closed form:
\begin{equation}
    q(\mathbf{z}_t|\mathbf{z}_0) = \mathcal{N}(\mathbf{z}_t; \sqrt{\bar{\alpha}_t}\mathbf{z}_0, (1-\bar{\alpha}_t)\mathbf{I}),
\end{equation}
where $\alpha_t = 1 - \beta_t$ and $\bar{\alpha}_t = \prod_{i=1}^{t} \alpha_i$.

The reverse process aims to recover the original latent vector $\mathbf{z}_0$ by iteratively denoising from a pure noise vector $\mathbf{z}_T \sim \mathcal{N}(\mathbf{0}, \mathbf{I})$. This is accomplished by a neural network, $\boldsymbol{\epsilon}_\theta(\mathbf{z}_t, t, \mathbf{c})$, which is trained to predict the noise component $\boldsymbol{\epsilon}$ from the noisy latent $\mathbf{z}_t$ at time step $t$, often under some conditioning $\mathbf{c}$ (e.g., text embeddings). The training objective is typically a simplified mean squared error loss:
\begin{equation}
\mathcal{L} = \mathbb{E}_{t, \mathbf{z}_0, \boldsymbol{\epsilon}} \left[ || \boldsymbol{\epsilon} - \boldsymbol{\epsilon}_\theta(\sqrt{\bar{\alpha}_t}\mathbf{z}_0 + \sqrt{1-\bar{\alpha}_t}\boldsymbol{\epsilon}, t, \mathbf{c}) ||^2 \right].
\end{equation}

For sampling, we employ the DDIM sampler, which offers a deterministic, faster sampling process. Given a noise prediction $\boldsymbol{\epsilon}_\theta(\mathbf{z}_t, t, \mathbf{c})$, the DDIM sampler first predicts the latent $\mathbf{\hat{z}}_0$ and then uses it to compute the denoised latent at the previous step, $\mathbf{z}_{t-1}$:
\begin{align}
 \mathbf{\hat{z}}_0 &= \frac{\mathbf{z}_t - \sqrt{1 - \bar{\alpha}_t} \boldsymbol{\epsilon}_\theta(\mathbf{z}_t, t, \mathbf{c})}{\sqrt{\bar{\alpha}_t}} \\
\mathbf{z}_{t-1} &= \sqrt{\bar{\alpha}_{t-1}}\mathbf{\hat{z}}_0 + \sqrt{1 - \bar{\alpha}_{t-1}} \boldsymbol{\epsilon}_\theta(\mathbf{z}_t, t, \mathbf{c}).
\label{eq:ddim}
\end{align}
This deterministic nature allows for consistent generation paths, which is a crucial property for caching mechanisms. We can thus define the DDIM step as an operator $\mathcal{D}(\mathbf{z}_t, \boldsymbol{\epsilon}_\theta, t)$, which computes $\mathbf{z}_{t-1}$ using the rule from Eq.~\ref{eq:ddim}.
\begin{equation}
\mathbf{z}_{t-1} = \mathcal{D}(\mathbf{z}_t, \boldsymbol{\epsilon}_\theta, t).
\label{eq:sampling}
\end{equation}

\begin{figure}[t!]
\centering
\includegraphics[width=1.0\linewidth]{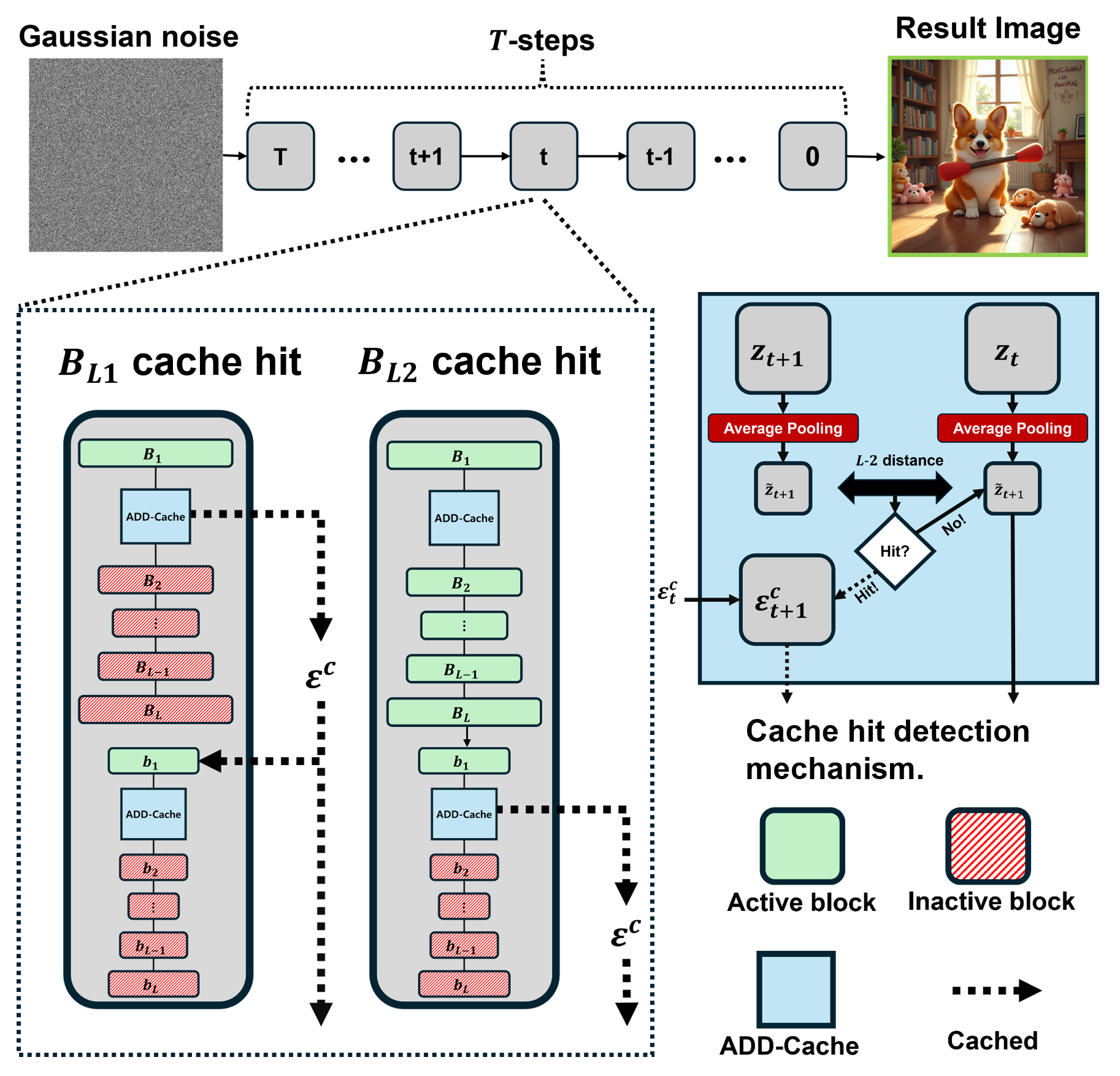}
\caption{Overview of the proposed H2-cache framework. The pipeline consists of a hierarchical, two-stage caching mechanism applied to the structure-defining ($\mathcal{B}_{L1}$) and detail-refining ($\mathcal{B}_{L2}$) stages. The right panel details the hit detection mechanism, which employs Pooled Feature Summarization for efficient similarity checks.}
\label{fig:proposed_method}
\end{figure}

\subsection{Denoising Network Architecture and Two-Stage Sampling}
In our work, we adopt the \textbf{Flux} \cite{blackforestlabs2024announcements,blackforestlabs2024flux} architecture as the denoising network $\boldsymbol{\epsilon}_\theta$. Unlike traditional U-Nets, Flux is a non-U-Net model composed of a series of transformer-based blocks. This choice is motivated by its impressive performance and architectural simplicity, making it an ideal candidate for acceleration.

A key feature of Flux is that its internal structure allows for the DDIM sampling step to be decomposed into a distinct two-stage process. Specifically, we decompose the single DDIM sampling step, previously defined as Eq.~\ref{eq:sampling}, into the following two-stage computational process. This is foundational to our proposed method, as illustrated in Fig.~\ref{fig:proposed_method}.

\paragraph{Step 1: Multi-Transformer Block Processing}
First, to capture the overall composition of the image, the operator $\mathcal{B}_{L1}$ processes the input latent $\mathbf{z}_t$ to produce an intermediate feature map $\mathbf{z}'_t$:
\begin{equation}
    \mathbf{z}'_t = \mathcal{B}_{L1}(\mathbf{z}_t, t, \mathbf{c}).
\end{equation}

\begin{figure*}
\centering
\includegraphics[width=1.0\linewidth]{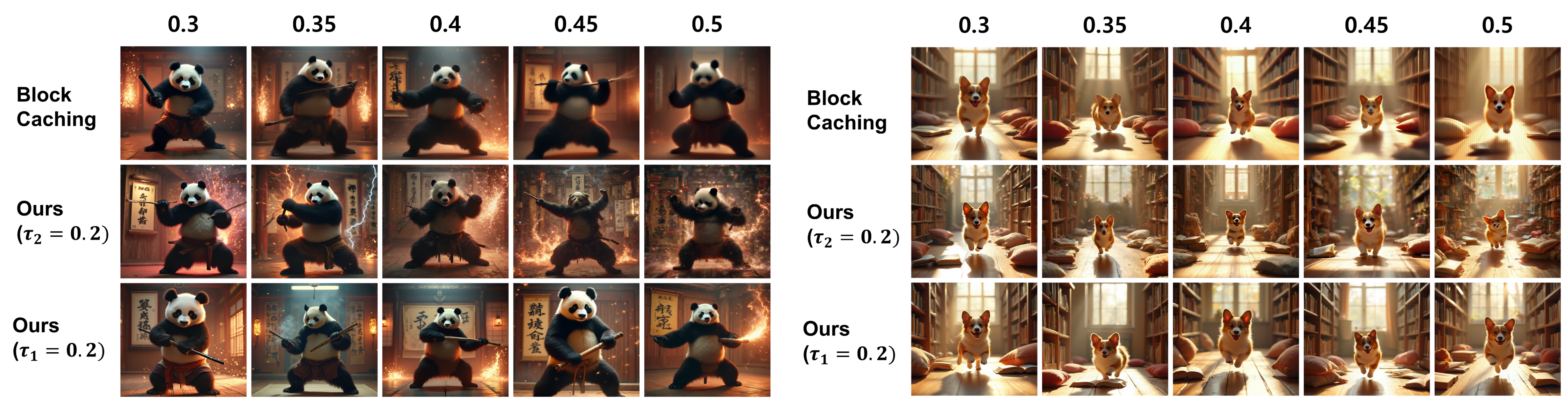}
\caption{Qualitative analysis of H2-cache. Caching the global structure stage, $\mathcal{B}_{L1}$ (bottom row), freezes the overall pose and layout. In contrast, caching the detail-refining stage, $\mathcal{B}_{L2}$ (middle row), preserves fine-grained textures while allowing the global structure to evolve, demonstrating a clear functional separation compared to standard block caching (top row).}
\label{fig:blockcaching}
\end{figure*}

\paragraph{Step 2: Single-Transformer Block Processing}
In the second stage, to generate the detailed parts of an image, the operator $\mathcal{B}_{L2}$ computes the noise prediction $\boldsymbol{\epsilon}_{\theta}$ from the intermediate features. This predicted noise is then immediately used to complete the single DDIM denoising step to obtain $\mathbf{z}_{t-1}$:
\begin{equation}
\mathbf{z}_{t-1} = \mathcal{D}\left(\mathbf{z}_{t}, \boldsymbol{\epsilon}_{\theta}=\mathcal{B}_{L2}(\mathbf{z'}_{t}, t, \mathbf{c}), t\right).
\end{equation}
This formulation explicitly shows the decomposition of a single sampling step into two functionally distinct computational stages ($\mathcal{B}_{L1}$ and $\mathcal{B}_{L2}$). As shown in Fig.~\ref{fig:blockcaching}, caching $\mathcal{B}_{L1}$ tends to freeze the global layout, whereas caching $\mathcal{B}_{L2}$ preserves fine details while allowing the structure to evolve, empirically validating this functional separation.

\section{Proposed Method}
\subsection{H2-cache: Hierarchical Two-Stage Caching}
The core idea of our method, H2-cache, is to exploit the functional separation within the denoising network's blocks. As established in our preliminaries and empirically validated in Fig.~\ref{fig:blockcaching}, the initial stage of a block, $\mathcal{B}_{L1}$, is primarily responsible for the image's global structure, while the subsequent stage, $\mathcal{B}_{L2}$, refines high-frequency details. Standard caching methods treat the entire block as a monolithic unit, failing to leverage this separation.

H2-cache introduces a hierarchical, two-stage caching mechanism that applies distinct caching logic to each stage. This enables more granular control over the speed-quality trade-off. The primary goal is to bypass the expensive $\mathcal{B}_{L1}$ computation when the image's structure has stabilized. However, even when the structure is still evolving (i.e., a $\mathcal{B}_{L1}$ cache miss), H2-cache provides a secondary opportunity to skip the $\mathcal{B}_{L2}$ computation if the details are sufficiently refined.

To implement this, our method maintains a cache storing a tuple from the last step where a full computation occurred: $\{\mathbf{z}_{\text{cache-in}}, \mathbf{z}'_{\text{cache}}, \boldsymbol{\epsilon}_{\text{cache}}\}$. The process for each timestep $t$, governed by thresholds $\tau_1$ and $\tau_2$, unfolds as follows (illustrated in Fig.~\ref{fig:proposed_method}):

\paragraph{Step 1: Joint Cache Check for Structure and Details.}
First, we check if the image has stabilized enough to reuse the entire cached result from a previous step. We compare the current input latent $\mathbf{z}_t$ with the input from the last fully computed step, $\mathbf{z}_{\text{cache-in}}$. A cache hit occurs if the L2 distance is below the primary threshold $\tau_1$.

If a cache hit occurs ($||\mathbf{z}_t - \mathbf{z}_{\text{cache-in}}||_2 < \tau_1$), it signifies high stability. We bypass both $\mathcal{B}_{L1}$ and $\mathcal{B}_{L2}$ computations entirely, reusing both the cached intermediate features and the final noise prediction.
\begin{align}
\mathbf{z}'_t &= \mathbf{z}'_{\text{cache}} \\
\boldsymbol{\epsilon}_\theta &= \boldsymbol{\epsilon}_{\text{cache}}
\end{align}
If this check fails (a cache miss), it indicates a structural change, and we proceed to compute $\mathcal{B}_{L1}$ and perform a secondary check.
\begin{equation}
\mathbf{z}'_t = \mathcal{B}_{L1}(\mathbf{z}_t, t, \mathbf{c})
\end{equation}

\paragraph{Step 2: Detail-Only Cache Check}
This step is executed only if the first check resulted in a miss. Having computed a new intermediate feature $\mathbf{z}'_t$, we now check if we can avoid computing the detail-refining block $\mathcal{B}_{L2}$. We compare the newly computed $\mathbf{z}'_t$ with its cached counterpart, $\mathbf{z}'_{\text{cache}}$.

If the L2 distance is below the secondary threshold $\tau_2$, we conclude that despite the structural shift, the resulting intermediate features are similar enough to reuse the cached detail computation. We thus reuse the final cached noise prediction $\boldsymbol{\epsilon}_{\text{cache}}$. Otherwise, we execute $\mathcal{B}_{L2}$ to generate a new noise prediction. The formal rule is:
\begin{equation}
\boldsymbol{\epsilon}_\theta = 
\begin{cases} 
    \boldsymbol{\epsilon}_{\text{cache}} & \text{if } ||\mathbf{z}'_t - \mathbf{z}'_{\text{cache}}||_2 < \tau_2 \\
    \mathcal{B}_{L2}(\mathbf{z}'_t, t, \mathbf{c}) & \text{otherwise}
\end{cases}
\label{eq:cache_b2}
\end{equation}

\paragraph{Step 3: Denoising Step and Cache Update.}
With the final noise prediction $\boldsymbol{\epsilon}_\theta$ determined (either from cache or computation), we compute the next denoised latent $\mathbf{z}_{t-1}$ using the DDIM operator:
\begin{equation}
    \mathbf{z}_{t-1} = \mathcal{D}(\mathbf{z}_t, \boldsymbol{\epsilon}_\theta, t).
\end{equation}
Finally, if any part of the block was recomputed (i.e., if Step 1 was a miss), we update the entire cache with the new values to ensure the cache holds the most recent fully computed results:
\begin{equation}
\text{if } ||\mathbf{z}_t - \mathbf{z}_{\text{cache-in}}||_2 \ge \tau_1: \quad
\begin{cases}
    \mathbf{z}_{\text{cache-in}} &\leftarrow \mathbf{z}_t \\
    \mathbf{z}'_{\text{cache}} &\leftarrow \mathbf{z}'_t \\
    \boldsymbol{\epsilon}_{\text{cache}} &\leftarrow \boldsymbol{\epsilon}_\theta
\end{cases}
\end{equation}

\subsection{Similarity Estimation via Pooled Feature Summarization}
Our H2-cache doubles the frequency of similarity checks compared to conventional methods, potentially performing this operation twice per block at every denoising step. This heightened computational demand renders a naive, full-tensor comparison intractable and makes efficiency a paramount concern. To mitigate this specific computational burden, we introduce a more sophisticated methodology: \textbf{pooled feature summarization (PFS)}. This technique provides a robust proxy for the true semantic difference between high-dimensional feature tensors while remaining computationally trivial.

The core principle of PFS is to determine the similarity between two tensors by first downsampling them into smaller summary tensors, or "thumbnails." This hit detection mechanism is visually summarized in Fig.~\ref{fig:proposed_method} (right). As illustrated, the process involves applying an average pooling operation to both the current and cached tensors, and then computing a relative difference metric on these compact representations to make a caching decision. The detailed mathematical formulation of this process is as follows.

Initially, the tensors are standardized to a 4-dimensional format, $\mathbf{T} \in \mathbb{R}^{B \times C \times H \times W}$. If an input tensor is not 4D, it is reshaped, with its channel dimension $C$ set to 1. The core of the downsampling process is controlled by a pre-defined integer divisor. Crucially, H2-Cache allows for setting this divisor independently for each of the two caching stages, enabling fine-tuned control over the similarity estimation. We define $D_{p1}$ for the structure-defining stage ($\mathcal{B}_{L1}$) and $D_{p2}$ for the detail-refining stage ($\mathcal{B}_{L2}$). This allows us, for example, to use a more aggressive summarization (larger $D_{p1}$) for the coarse structural check and a more conservative one (smaller $D_{p2}$) for the finer detail check. The kernel size and stride for the pooling operation, $S_k$, is then calculated based on the tensor's height $H$ and the corresponding divisor for the current stage, $D_{pi}$ (where $i \in \{1, 2\}$):
\begin{equation}
S_k = \lfloor H / D_{pi} \rfloor
\end{equation}
This single value $S_k$ is then used to define a symmetric 2D average pooling operation that is applied to both tensors to produce their respective thumbnails, $\tilde{\mathbf{T}}_1$ and $\tilde{\mathbf{T}}_2$. The operation is:
\begin{equation}
    \tilde{\mathbf{T}} = \text{AvgPool}_{\text{kernel}=(S_k, S_k), \text{stride}=(S_k, S_k)}(\mathbf{T})
\end{equation}
This transformation yields a thumbnail tensor $\tilde{\mathbf{T}} \in \mathbb{R}^{B \times C \times H' \times W'}$, where the new spatial dimensions $H'$ and $W'$ are calculated as:
\begin{align}
    H' &= \lfloor (H - S_k) / S_k \rfloor + 1 \\
    W' &= \lfloor (W - S_k) / S_k \rfloor + 1
\end{align}
Finally, we compute the relative difference metric $D$ on these compact representations. The difference between the current tensor $\mathbf{T}^{(t)}$ and the cached tensor $\mathbf{T}^{(t-1)}$ is computed as:
\begin{equation}
    D(\mathbf{T}^{(t)}, \mathbf{T}^{(t-1)}) = \frac{\mathbb{E}\left[|\tilde{\mathbf{T}}^{(t)} - \tilde{\mathbf{T}}^{(t-1)}|\right]}{\mathbb{E}\left[|\tilde{\mathbf{T}}^{(t-1)}|\right]},
\end{equation}
where the expectation $\mathbb{E}[\cdot]$ is calculated over all elements of the summarized tensors. This approach offers significant advantages. The use of hardware-accelerated average pooling is exceptionally fast, and by averaging over local patches, the resulting thumbnail is less sensitive to high-frequency noise, yielding a more stable signal for the caching decision.

\section{Evaluation}
\subsection{Experiment Setup}
All experiments were conducted using nunchaku \cite{li2025svdquant}, a lightweight framework engine designed for efficient model execution. This ensured a consistent and controlled environment for evaluating the performance of our proposed method. Unless otherwise specified, all image generation experiments were performed using the following default parameters: a resolution of 1024x1024 pixels, a guidance scale of 3.5, and a batch of 1. The denoising process was carried out over 100 inference steps. For the baseline denoising network, we employed the flux.1-dev \footnote{\url{mit-han-lab/nunchaku-flux.1-dev/svdq-int4_r32-flux.1-dev}} model, a quantized version of the Flux architecture optimized for high-throughput generation. The hardware configuration for all evaluations consisted of a single NVIDIA A5000 GPU and an Intel(R) Core(TM) i9-14900K CPU. For quantitative evaluation, we generated images using prompts extracted from the CUTE80 dataset via the LLaVA-NeXT-8B model. The generation pipeline utilized a transformer-based model enhanced with Low-Rank Adaptation (LoRA) to improve visual realism. We integrated the \texttt{XLabs-AI/flux-RealismLora}  \footnote{\url{https://huggingface.co/XLabs-AI/flux-RealismLora}} weights into the transformer and set the LoRA strength to 0.8.

\subsection{Experiment results}

\begin{figure*}
\centering
\includegraphics[width=1.0\linewidth]{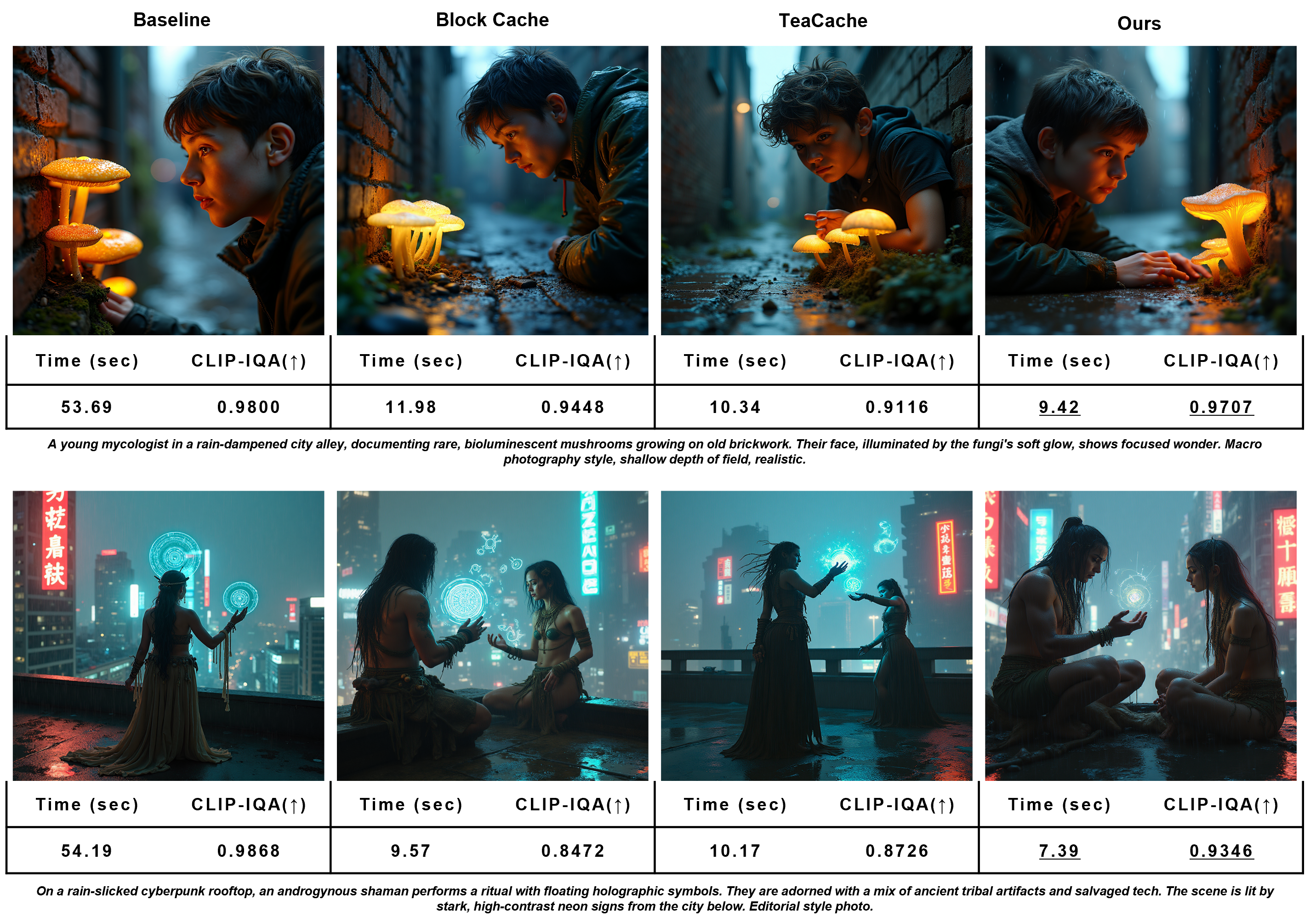}
\caption{Our method (H2-cache) is compared against three baselines: no caching (Baseline), block cache, and TeaCache. For each method, we report the inference time and the reference-free CLIP-IQA($\uparrow$) score \cite{wang2023exploring}.}
\label{exper:main_exper}
\end{figure*}
The primary qualitative and quantitative results of our method are presented in Fig.~\ref{exper:main_exper}. We compare our proposed H2-cache against three key baselines: the standard denoising process without caching (Baseline), block cache \cite{cheng2025paraattention}, and TeaCache \cite{liu2025timestep}. The results clearly demonstrate the effectiveness of our approach. The baseline method, without any caching, requires over 53 seconds for generation. In contrast, all caching methods provide a significant speedup, reducing the inference time to under 12 seconds. However, the trade-off in image quality is stark for existing methods. For the first prompt, block Cache and TeaCache cause the CLIP-IQA score to drop from the baseline's 0.9800 to 0.9448 and 0.9116, respectively. However, our method achieves the fastest inference time of 9.42 s while maintaining a CLIP-IQA score of 0.9707, which is remarkably close to the baseline. A similar trend is observed for the second prompt, where our method achieves a 7.33x speedup (54.19 to 7.39 s) with only a minor drop in quality (0.9868 to 0.9346), far outperforming the other caching techniques. Our method successfully preserves the intricate details and semantic concepts of the prompt confirming its ability to accelerate inference without sacrificing perceptual quality.

\begin{table}[h!]
\centering
\caption{Quantitative comparison on the CUTE80 dataset. Values in parentheses denote the speedup factor (for Time) and percentage change (for CLIP-IQA) relative to the Baseline. \goodchange{Red} indicates an improvement, while \badchange{blue} indicates a regression.}
\label{tab:cute80_comparison_final}
\begin{tabular}{l c c c}
\toprule
& \multicolumn{2}{c}{\textbf{Time (s)} $\downarrow$} & {\textbf{CLIP-IQA} $\uparrow$} \\
\cmidrule(lr){2-3} \cmidrule(lr){4-4}
\textbf{Method} & \textbf{Avg. (Speedup)} & \textbf{Std.} & \textbf{Avg. (Change \%)} \\
\midrule
Baseline & 55.72 (1.0$\times$) & 0.14 & 0.7693 (-) \\
Block Cache & 12.82 (\goodchange{4.35$\times$}) & 0.41 & 0.7681 (\badchange{-0.16\%}) \\
teaCache & 11.16 (\goodchange{4.99$\times$}) & 0.14 & 0.7462 (\badchange{-3.00\%}) \\
\textbf{Ours} & \textbf{10.97 (\goodchange{5.08$\times$})} & \textbf{0.36} & \textbf{0.7688 (\badchange{-0.07\%})} \\
\bottomrule
\end{tabular}
\end{table}
Table~\ref{tab:cute80_comparison_final} presents a quantitative comparison of our method against several baselines on the CUTE80 dataset, evaluating both processing time and image quality using CLIP-IQA. For the caching methods, the thresholds were set to $\tau=1.0$ for Block Cache and $\tau=1.0$ for teaCache. The results indicate that while the Baseline method achieves the highest CLIP-IQA score of 0.7693, it is substantially slower, with an average time of 55.72 seconds. In contrast, our proposed method demonstrates a highly efficient profile. It reduces the processing time dramatically to 10.97 seconds, achieving the fastest performance among all methods. Crucially, unlike TeaCache which suffers a significant 3\% drop in quality (0.7462 CLIP-IQA), our method maintains a high-quality score of 0.7688. This score represents a negligible regression of only -0.07\% from the baseline, making it virtually identical in perceptual quality. Therefore, our approach provides the most effective balance, achieving a 5.08$\times$ speedup over the baseline while preserving image quality, and outperforming other caching methods in the overall trade-off between speed and quality.

\subsection{Ablation Study}
\begin{figure}[h!]
\centering
\includegraphics[width=1.0\linewidth]{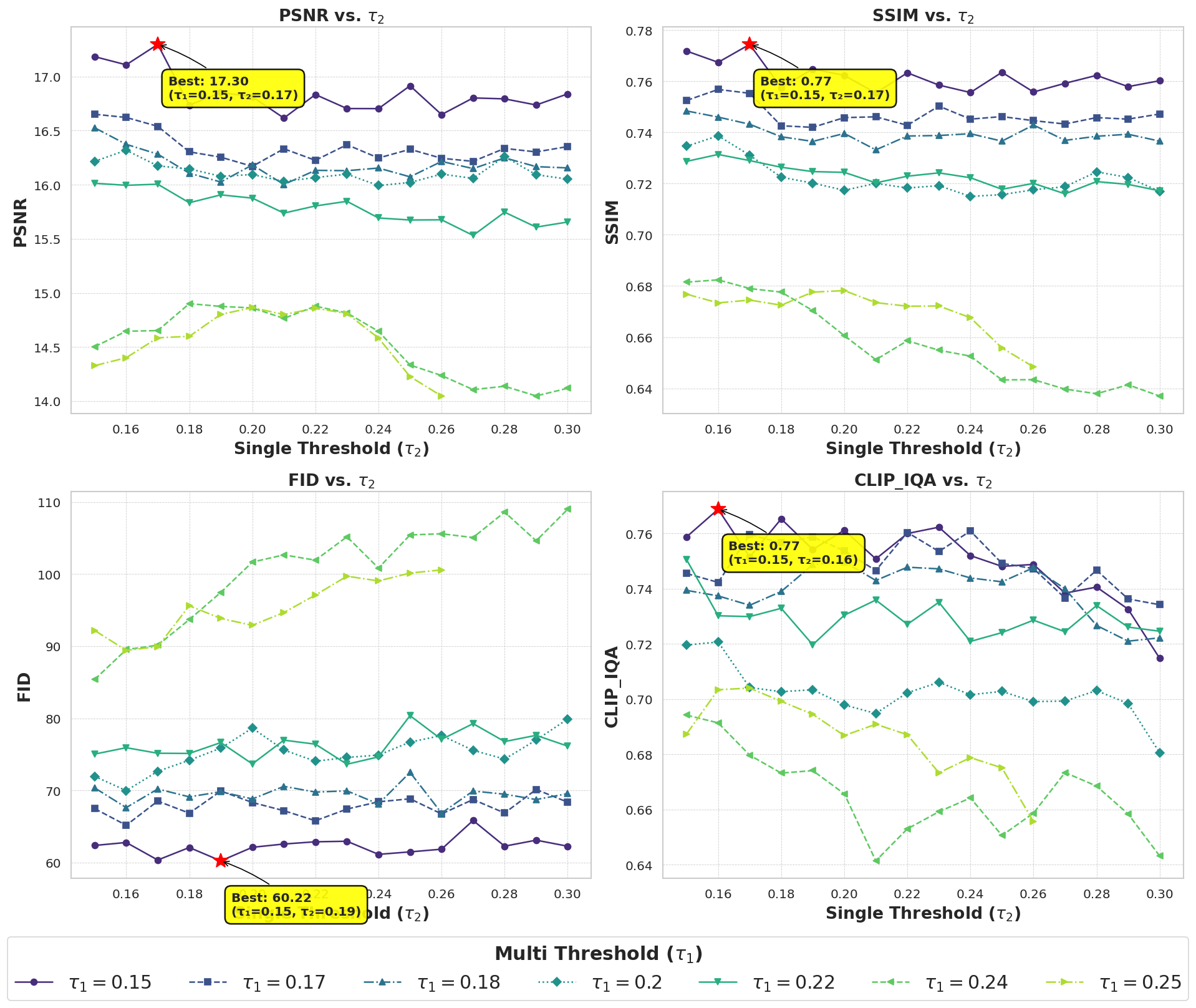}
\caption{Quantitative evaluation of performance across various image quality metrics (PSNR, SSIM, FID, and CLIP-IQA) by varying the structural threshold ($\tau_1$) and the detail threshold ($\tau_2$). The red stars indicate the best-performing hyperparameter set for each metric.}
\label{exper:threshold}
\end{figure}

Fig.~\ref{exper:threshold} presents a quantitative analysis of our method by evaluating the trade-offs between the hierarchical thresholds, $\tau_1$ (structural) and $\tau_2$ (detail). We plot four standard image quality metrics—PSNR, SSIM, FID, and CLIP-IQA—as a function of $\tau_2$ for several fixed values of $\tau_1$. The results reveal a complex interplay between these parameters. Notably, a low structural threshold of $\tau_1=0.15$ (solid dark blue line) consistently achieves the best peak performance across all metrics, as highlighted by the red stars: a PSNR of 17.30 and SSIM of 0.77 are achieved at $\tau_2=0.17$, a CLIP-IQA score of 0.77 at $\tau_2=0.16$, and a superior FID score of 60.22 (where lower is better) at $\tau_2=0.19$. However, this top performance is highly sensitive to the choice of $\tau_2$, exhibiting sharp peaks. Conversely, higher values for $\tau_1$ generally lead to more stable but sub-optimal performance across the range of $\tau_2$. This analysis demonstrates that there is no single, universally optimal setting; instead, the ideal configuration depends on the desired balance between peak performance and stability for a given metric. It empirically validates our hierarchical design, showing that granular control over both structural and detail caching stages is crucial for optimizing the generation process.

\begin{table*}
\centering
\caption{Performance impact of enabling Pooled Feature Summarization (PFS). The table compares our method with PFS against a baseline version without PFS (w/o PFS) for a fixed $\tau_{1}=0.15$. All percentage changes in parentheses are relative to the w/o PFS baseline. \textcolor{red}{Red} indicates improvement, while \textcolor{blue}{blue} indicates degradation.}
\label{tab:full_performance_delta}
\resizebox{\textwidth}{!}{%
\begin{tabular}{l lllll}
\toprule
$\tau_{2}$ & \textbf{Time (s)} & \textbf{PSNR} & \textbf{SSIM} & \textbf{FID} & \textbf{CLIP-IQA} \\
\midrule
0.15 & 11.30 $\pm$ 0.33 (\textcolor{red}{-14.5\%}) & 17.18 $\pm$ 4.22 (\textcolor{red}{+1.0\%}) & 0.77 $\pm$ 0.12 (\textcolor{red}{+0.7\%}) & 62.56 $\pm$ 0.16 (\textcolor{red}{-1.1\%}) & 0.76 $\pm$ 0.17 (\textcolor{blue}{-0.4\%}) \\
0.18 & 10.97 $\pm$ 0.36 (\textcolor{red}{-5.0\%}) & 16.74 $\pm$ 4.23 (\textcolor{blue}{-3.1\%}) & 0.76 $\pm$ 0.12 (\textcolor{blue}{-2.3\%}) & 62.08 $\pm$ 0.16 (\textcolor{blue}{+2.2\%}) & 0.77 $\pm$ 0.16 (\textcolor{red}{+1.7\%}) \\
0.20 & 10.93 $\pm$ 0.29 (\textcolor{red}{-4.0\%}) & 16.81 $\pm$ 4.18 (\textcolor{blue}{-1.8\%}) & 0.76 $\pm$ 0.12 (\textcolor{blue}{-1.0\%}) & 62.10 $\pm$ 0.15 (\textcolor{blue}{+3.5\%}) & 0.76 $\pm$ 0.15 (\textcolor{red}{+0.2\%}) \\
0.22 & 10.89 $\pm$ 0.28 (\textcolor{red}{-1.9\%}) & 16.89 $\pm$ 4.00 (\textcolor{blue}{-1.1\%}) & 0.76 $\pm$ 0.12 (\textcolor{blue}{-0.7\%}) & 62.87 $\pm$ 0.17 (\textcolor{blue}{+9.4\%}) & 0.77 $\pm$ 0.15 (\textcolor{red}{+0.8\%}) \\
0.24 & 10.76 $\pm$ 0.24 (\textcolor{red}{-3.0\%}) & 16.70 $\pm$ 4.18 (\textcolor{blue}{-0.5\%}) & 0.76 $\pm$ 0.12 (\textcolor{blue}{-0.5\%}) & 61.13 $\pm$ 0.16 (\textcolor{red}{-3.8\%}) & 0.75 $\pm$ 0.16 (\textcolor{blue}{-1.9\%}) \\
0.25 & 10.67 $\pm$ 0.27 (\textcolor{red}{-3.8\%}) & 16.91 $\pm$ 4.08 (\textcolor{blue}{-0.1\%}) & 0.76 $\pm$ 0.11 (\textcolor{blue}{-1.3\%}) & 61.46 $\pm$ 0.15 (\textcolor{red}{-1.6\%}) & 0.75 $\pm$ 0.17 (\textcolor{blue}{-2.6\%}) \\
0.27 & 10.44 $\pm$ 0.28 (\textcolor{red}{-5.8\%}) & 16.80 $\pm$ 4.21 (\textcolor{blue}{-0.6\%}) & 0.76 $\pm$ 0.12 (\textcolor{blue}{-0.7\%}) & 65.82 $\pm$ 0.17 (\textcolor{blue}{+9.6\%}) & 0.74 $\pm$ 0.17 (\textcolor{blue}{-3.7\%}) \\
0.29 & 10.30 $\pm$ 0.28 (\textcolor{red}{-6.9\%}) & 16.74 $\pm$ 4.18 (\textcolor{blue}{-0.3\%}) & 0.76 $\pm$ 0.12 (\textcolor{blue}{-0.4\%}) & 63.93 $\pm$ 0.17 (\textcolor{blue}{+1.6\%}) & 0.73 $\pm$ 0.17 (\textcolor{blue}{-3.4\%}) \\
0.30 & 10.17 $\pm$ 0.29 (\textcolor{red}{-8.2\%}) & 16.94 $\pm$ 4.13 (\textcolor{blue}{-0.1\%}) & 0.76 $\pm$ 0.11 (\textcolor{blue}{-0.5\%}) & 62.25 $\pm$ 0.18 (\textcolor{red}{-1.2\%}) & 0.73 $\pm$ 0.18 (\textcolor{blue}{-3.9\%}) \\
\bottomrule
\end{tabular}
}
\end{table*}
Table~\ref{tab:full_performance_delta} presents a detailed ablation study isolating the impact of our Pooled Feature Summarization (PFS) technique. The table compares our full method (with PFS) against a version without it (w/o PFS) across various $\tau_{2}$ settings. The results clearly demonstrate that integrating PFS consistently yields a notable acceleration, reducing processing time by up to 14.5\%. While this optimization introduces a slight trade-off, its impact on image quality is not significant. The degradation in key metrics such as PSNR and SSIM is generally marginal, with most changes remaining below 3\%. This analysis validates our claim that PFS is a powerful and efficient optimization, providing a significant enhancement in computational efficiency while maintaining a high degree of image quality with minimal and acceptable losses.

\begin{figure}
\centering
\includegraphics[width=1.0\linewidth]{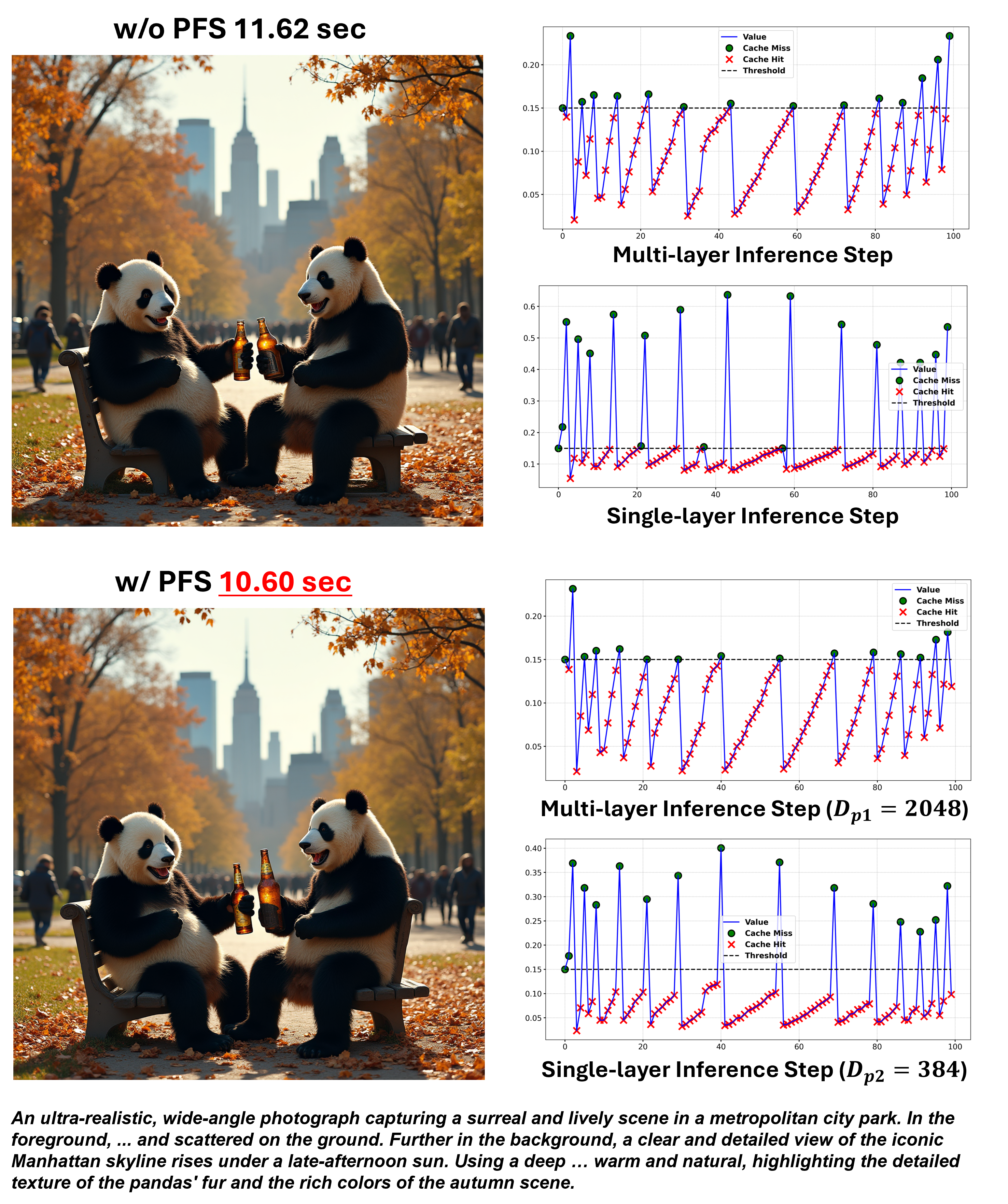}
\caption{Qualitative comparison of caching behavior with and without PFS.}
\label{exper:pfs_comparison}
\end{figure}
Fig.~\ref{exper:pfs_comparison} illustrates the practical impact of our PFS on the caching process during image generation. The top row (w/o PFS) depicts a baseline approach where caching decisions rely on a direct comparison metric governed by fixed thresholds, such as $\tau_{1}=0.15$ and $\tau_2=0.20$. The corresponding graphs show that the calculated difference value fluctuates significantly, leading to a specific pattern of cache hits and misses. In contrast, the bottom row (w/ PFS) demonstrates our proposed method, which integrates PFS with varying summary sizes (e.g., $D_{p1}=512, D_{p2}=384$) alongside the thresholds. The resulting difference metric, plotted over the inference steps, is visibly more stable and consistent. This stability allows for more effective and reliable caching decisions, which translates directly to a reduction in the total generation time, decreasing from 12.11 to 11.15 s in one case and from 10.22 to 9.78 s in the other. This experiment validates that PFS serves as a computationally efficient and robust proxy for tensor similarity, making our frequent-checking cache strategy both feasible and effective.

\begin{table}
\centering
\caption{Performance of H2-Cache relative to the no-cache baseline. Average values are reported with their standard deviation ($\pm$ std). Values in parentheses denote the speedup factor and percentage change. \goodchange{Red} indicates improvement, while \badchange{blue} indicates regression.}
\label{tab:h2cache_final_performance}
\begin{tabular}{l c c}
\toprule
\textbf{Step Size} & \textbf{Time (s)} $\downarrow$ & \textbf{CLIP-IQA} $\uparrow$ \\
\midrule
\textbf{10}  & 4.89 $\pm$ 0.11 (\goodchange{1.22$\times$})  & 0.45 (\badchange{-13.74\%}) \\
\textbf{30}  & 7.42 $\pm$ 0.25 (\goodchange{2.29$\times$})  & 0.76 (\badchange{-3.38\%})  \\
\textbf{50}  & 8.59 $\pm$ 0.34 (\goodchange{3.26$\times$})  & 0.75 (\badchange{-4.67\%})  \\
\textbf{70}  & 9.37 $\pm$ 0.31 (\goodchange{4.17$\times$})  & 0.77 (\badchange{-0.32\%})   \\
\textbf{100} & 10.97 $\pm$ 0.36 (\goodchange{5.08$\times$}) & 0.77 (\badchange{-0.07\%})  \\
\bottomrule
\end{tabular}
\end{table}
Table~\ref{tab:h2cache_final_performance} presents the quantitative analysis of H2-Cache, demonstrating its performance gains over a standard inference process with no caching enabled (no-cache baseline). For this comparison, the H2-Cache thresholds were set to $\tau_{1}=0.15$ and $\tau_{2}=0.18$. The results confirm that our method achieves substantial acceleration, with the speed-up factor scaling impressively from 1.22$\times$ at 10 steps to over 5$\times$ at 100 steps. This trend highlights that the method is more effective at larger step sizes; the computational savings from skipping blocks increasingly outweigh the minor overhead introduced by the cache-hit detection mechanism. This significant reduction in generation time is achieved with a minimal and acceptable trade-off in the CLIP-IQA score. Notably, at 100 steps, our method provides a 5.08$\times$ speed-up while the image quality remains almost identical to the baseline, with a negligible regression of only -0.07\%. This demonstrates that H2-Cache is a highly effective and robust solution for accelerating diffusion model inference while preserving high perceptual quality.

\section{Conclusion}
In this work, we introduced H2-cache, a novel mechanism that successfully resolves the critical trade-off between inference speed and perceptual quality in modern diffusion models. We demonstrated that by identifying and exploiting the functional separation between a structure-defining stage and a detail-refining stage, it is possible to design a far more intelligent caching strategy. Our hierarchical, dual-threshold approach effectively implements this insight, aggressively optimizing the initial, structurally formative computations while protecting the latter, detail-oriented refinement steps. The experimental results confirmed our hypothesis, showing that H2-cache achieves up to a 5.08x speedup with negligible degradation in image quality, thereby significantly outperforming existing methods. While our method proves effective, its current implementation relies on empirically set thresholds ($\tau_1, \tau_2$). A promising direction for future work is the development of a learnable policy to automate the selection of these parameters, further enhancing adaptability. Furthermore, extending the core principles of H2-cache to accelerate other generative modalities, such as video and 3D diffusion models, presents an exciting avenue for research. Ultimately, H2-cache provides a robust and practical solution that makes high-fidelity generative AI more efficient and accessible for real-world applications.

% \section*{Acknowledgments}
% This work was supported by the National Research Foundation of Korea(NRF) grant funded by the Korea government(MSIT) (RS-2025-00559998 and RS-2023-0027116).

%Bibliography
\bibliographystyle{unsrt}  
\bibliography{references}

\begin{thebibliography}{10}

\bibitem{sohl2015deep}
Jascha Sohl‐Dickstein, Eric Weiss, Niru Maheswaranathan, and Surya Ganguli.
\newblock Deep unsupervised learning using nonequilibrium thermodynamics.
\newblock In {\em Proc. 32nd International Conference on Machine Learning}, pages 2256--2265, Jul 2015.

\bibitem{song2021score}
Yang Song, Jascha Sohl‐Dickstein, Diederik~P. Kingma, Abhishek Kumar, Stefano Ermon, and Ben Poole.
\newblock Score-based generative modeling through stochastic differential equations.
\newblock In {\em Proc. 9th International Conference on Learning Representations}, May 2021.

\bibitem{croitoru2023diffusion}
Florinel-Alin Croitoru, Vlad Hondru, Radu~Tudor Ionescu, and Mubarak Shah.
\newblock Diffusion models in vision: A survey.
\newblock {\em IEEE Transactions on Pattern Analysis and Machine Intelligence}, 45(9):10850--10869, Sep 2023.

\bibitem{ha2025generative}
Sangjun Ha, Mingyu Sung, Faisal Saeed, Sangseok Yun, Il-Min Kim, and Jae-Mo Kang.
\newblock Generative-diffusion-model-based deep-learning framework for remaining useful life prediction.
\newblock {\em IEEE Internet of Things Journal}, 12(11):18431--18434, Jun 2025.

\bibitem{saharia2022photorealistic}
Chitwan Saharia, William Chan, Saurabh Saxena, Lala Li, Jay Whang, Emily~L. Denton, Kamyar Ghasemipour, Raphael Gontijo~Lopes, Burcu Karagol~Ayan, Tim Salimans, et~al.
\newblock Photorealistic text-to-image diffusion models with deep language understanding.
\newblock In {\em Proc. 36th International Conference on Neural Information Processing Systems}, volume~35, pages 36479--36494, Dec 2022.

\bibitem{ho2020denoising}
Jonathan Ho, Ajay Jain, and Pieter Abbeel.
\newblock Denoising diffusion probabilistic models.
\newblock In {\em Proc. 33rd International Conference on Neural Information Processing Systems}, volume~33, pages 6840--6851, Dec 2020.

\bibitem{song2021denoisingimplicit}
Jiaming Song, Chenlin Meng, and Stefano Ermon.
\newblock Denoising diffusion implicit models.
\newblock In {\em Proc.\ 9th International Conference on Learning Representations (ICLR)}, May 2021.

\bibitem{salimans2022progressive}
Tim Salimans and Jonathan Ho.
\newblock Progressive distillation for fast sampling of diffusion models.
\newblock {\em arXiv preprint arXiv:2202.00512}, 2022.

\bibitem{karras2022elucidating}
Tero Karras, Miika Aittala, Timo Aila, and Samuli Laine.
\newblock Elucidating the design space of diffusion-based generative models.
\newblock In {\em Proc. 36th International Conference on Neural Information Processing Systems}, volume~35, pages 26565--26577, Dec 2022.

\bibitem{rombach2022high}
Robin Rombach, Andreas Blattmann, Dominik Lorenz, Patrick Esser, and Bj{\"o}rn Ommer.
\newblock High-resolution image synthesis with latent diffusion models.
\newblock In {\em Proc. IEEE/CVF Conf. on Computer Vision and Pattern Recognition}, pages 10684--10695, Jun 2022.

\bibitem{lyu2022accelerating}
Zhaoyang Lyu, Xudong Xu, Ceyuan Yang, Dahua Lin, and Bo~Dai.
\newblock Accelerating diffusion models via early stop of the diffusion process.
\newblock {\em arXiv preprint arXiv:2205.12524}, May 2022.

\bibitem{wimbauer2024cache}
Felix Wimbauer, Bichen Wu, Edgar Schoenfeld, Xiaoliang Dai, Ji~Hou, Zijian He, Artsiom Sanakoyeu, Peizhao Zhang, Sam Tsai, Jonas Kohler, et~al.
\newblock Cache me if you can: Accelerating diffusion models through block caching.
\newblock In {\em Proc. IEEE/CVF Conf. on Computer Vision and Pattern Recognition}, pages 6211--6220, Jun 2024.

\bibitem{nichol2022glide}
Alexander~Quinn Nichol, Prafulla Dhariwal, Aditya Ramesh, Pranav Shyam, Pamela Mishkin, Bob McGrew, Ilya Sutskever, and Mark Chen.
\newblock Glide: Towards photorealistic image generation and editing with text-guided diffusion models.
\newblock In {\em Proc. 39th International Conference on Machine Learning}, volume 162, pages 16784--16804. PMLR, Jul 2022.

\bibitem{ramesh2022hierarchical}
Aditya Ramesh, Prafulla Dhariwal, Alex Nichol, Casey Chu, and Mark Chen.
\newblock Hierarchical text-conditional image generation with clip latents.
\newblock {\em arXiv preprint arXiv:2204.06125}, Apr 2022.

\bibitem{radford2021learningtransferablevisualmodels}
Alec Radford, Jong~Wook Kim, Chris Hallacy, Aditya Ramesh, Gabriel Goh, Sandhini Agarwal, Girish Sastry, Amanda Askell, Pamela Mishkin, Jack Clark, Gretchen Krueger, and Ilya Sutskever.
\newblock Learning transferable visual models from natural language supervision.
\newblock In {\em Proc. 38th International Conference on Machine Learning}, volume 139, pages 8748--8763. PMLR, Jul 2021.

\bibitem{liu2025timestep}
Feng Liu, Shiwei Zhang, Xiaofeng Wang, Yujie Wei, Haonan Qiu, Yuzhong Zhao, Yingya Zhang, Qixiang Ye, and Fang Wan.
\newblock Timestep embedding tells: It's time to cache for video diffusion model.
\newblock In {\em Proc. IEEE/CVF Conf. on Computer Vision and Pattern Recognition}, pages 7353--7363, Jun 2025.

\bibitem{blackforestlabs2024announcements}
{Black Forest Labs}.
\newblock Announcements.
\newblock \url{https://blackforestlabs.ai/announcements/}, 2024.
\newblock Accessed: 2025-07-11.

\bibitem{blackforestlabs2024flux}
{Black Forest Labs}.
\newblock Flux.
\newblock \url{https://github.com/black-forest-labs/flux}, Feb 2024.
\newblock Accessed: 2025-07-11.

\bibitem{li2025svdquant}
Muyang Li, Yujun Lin, Zhekai Zhang, Tianle Cai, Xiuyu Li, Junxian Guo, Enze Xie, Chenlin Meng, Jun-Yan Zhu, and Song Han.
\newblock Svdquant: Absorbing outliers by low-rank components for 4-bit diffusion models.
\newblock In {\em Proc. 13th International Conference on Learning Representations}, May 2025.

\bibitem{wang2023exploring}
Jianyi Wang, Kelvin C.~K. Chan, and Chen~Change Loy.
\newblock Exploring clip for assessing the look and feel of images.
\newblock In {\em Proc. AAAI Conf. on Artificial Intelligence}, volume~37, pages 2555--2563, Jun 2023.

\bibitem{cheng2025paraattention}
Zeyi Cheng.
\newblock Paraattention.
\newblock \url{https://github.com/chengzeyi/ParaAttention}, 2025.
\newblock Accessed: 2025-07-11.

\end{thebibliography}

\end{document}